%% file: main.tex
\documentclass[preprint,12pt]{elsarticle}

\usepackage{amssymb}
\usepackage{amsmath}
\usepackage{bm}
\usepackage{subfig}
\usepackage{wrapfig}
\usepackage{lipsum}     
\usepackage{float} 
\usepackage{graphicx}           
\usepackage{url} 
\usepackage{soul}
\usepackage{enumitem}
\usepackage{diagbox}
\usepackage{color}

\journal{Journal of Computational Physics}

\begin{document}

\begin{frontmatter}


\title{Adversarial Uncertainty Quantification in Physics-Informed Neural Networks}



\author{Yibo Yang}
\author{Paris Perdikaris}

\address{Department of Mechanical Engineering and Applied Mechanics, \\
University of Pennsylvania, \\
Philadelphia, PA 19\textcolor{black}{1}04, USA}

\begin{abstract}
We present a deep learning framework for quantifying and propagating uncertainty in systems governed by non-linear differential equations using physics-informed neural networks. Specifically, we employ latent variable models to construct probabilistic representations for the system states, and put forth an adversarial inference procedure for training them on data, while constraining their predictions to satisfy given physical laws expressed by partial differential equations. Such physics-informed constraints provide a regularization mechanism for effectively training deep generative models as surrogates of physical systems in which the cost of data acquisition is high, and training data-sets are typically small. This provides a flexible framework for characterizing  uncertainty in the outputs of physical systems due to randomness in their inputs or noise in their observations that entirely bypasses the need for repeatedly sampling expensive experiments or numerical simulators. We demonstrate the effectiveness of our approach through a series of examples involving uncertainty propagation in non-linear conservation laws, and the discovery of constitutive laws for flow through porous media directly from noisy data.
\end{abstract}

\begin{keyword}
Variational inference \sep Generative adversarial networks \sep Probabilistic deep learning \sep Probabilistic scientific computing \sep Data-driven modeling


\end{keyword}

\end{frontmatter}


\section{Introduction}\label{sec:intro}
Recent advances in machine learning and data analytics have yielded transformative results across diverse scientific disciplines, including image recognition \cite{krizhevsky2012imagenet}, natural language processing \cite{lecun2015deep}, cognitive science \cite{lake2015human}, and genomics \cite{alipanahi2015predicting}.
In all aforementioned areas, the volume of data has increased substantially compared to even a decade ago, but analyzing big data is expensive and time-consuming. Data-driven methods, which have been enabled  by the availability of sensors, data storage, and computational resources, are taking center stage across many disciplines of science. We now have highly scalable solutions for problems in object detection and recognition, machine translation, text-to-speech conversion, recommender systems, and information retrieval \cite{lecun2015deep}. All of these solutions attain state-of-the-art performance when trained with large amounts of data. 

However, more often than not, in laboratory experiments and large-scale simulations aiming to elucidate and predict complex phenomena, a large number of quality and error-free data is prohibitively costly to obtain. Under this setting, purely data-driven approaches for machine learning present difficulties when the data is scarce relative to the complexity of the system. The vast majority of state-of-the art machine learning techniques (e.g., deep neural nets, convolutional networks, recurrent networks, etc. \cite{goodfellow2016deep}) are lacking robustness and fail to provide any guarantees of convergence or quantify the error/uncertainty associated with their predictions. Hence, the ability to learn in a robust and sample-efficient manner is a necessity in these data-limited domains. Even less well understood is how one can constrain such algorithms to leverage domain-specific knowledge and return predictions that satisfy certain physical principles (e.g., conservation of mass, momentum, etc.). 

These shortcomings often generate skepticism and disbelief among applied mathematicians and engineers regarding the solid grounding of purely data-driven machine learning approaches. In recent work, Raissi {\em et. al.} \cite{raissi2017physics1,raissi2017physics2,raissi2018multistep,raissi2018hidden,raissi2018deep} set foot exactly at this relatively unexplored interface between applied mathematics and contemporary machine learning by revisiting the idea of penalizing the loss function of deep neural networks using differential equation constraints, as first put forth by Psichogios and Ungar \cite{psichogios1992hybrid} and Lagaris {\em et. al.} \cite{lagaris1998artificial}.
This line of work has empirically demonstrated how such physics-informed constraints regularize learning in {\em small data} regimes, can lead to the discovery of governing equations and reduced-order models, as well as enable the prediction of complex dynamics from incomplete models and incomplete data. Despite a series of impressive results in canonical problems, Raissi {\em et. al.} \cite{raissi2017physics1,raissi2017physics2} have also pointed out cases in which the training phase of these algorithms faces severe difficulties for reasons that are currently poorly understood. In lack of supporting theory on convergence and a-posteriori error estimation, this naturally poses the need for   for scalable algorithms for uncertainty quantification.

A literature review of the current state-of-the-art in uncertainty quantification reveals a subtle dichotomy between different communities. On one hand, researchers in applied mathematics and scientific computing predominately rely on mathematical models that are rigorously derived from first physical principles. At the dawn of exascale computing, such models have enabled the accurate simulation of increasingly more complex phenomena (see for e.g., \cite{perdikaris2016multiscale,rossinelli2015silico}). They have also enabled {\em in-silico} systematic studies in which the behavior of a system can be probed in a controlled fashion for different conditions, parameter settings, external inputs, etc. \cite{vsukys2017optimal}. The latter aims to both elucidate the key mechanisms that govern the behavior of a system, but also characterize the robustness of the resulting predictions with respect to epistemic and aleatory uncertainty \cite{oden2010computer}. However, despite the fact that much progress has been made over the last two decades, the most popular  methods for scientific computing under uncertainty, such as polynomial chaos expansions \cite{ghanem1990polynomial, xiu2002wiener, najm2009uncertainty}, sparse grid quadratures \cite{gerstner1998numerical, eldred2009comparison}, multi-level/multi-fidelity Monte Carlo sampling \cite{barth2011multi, peherstorfer2016optimal}, proper orthogonal decomposition \cite{berkooz1993proper, le2010spectral}, and Gaussian process regression models \cite{bilionis2012multi, bilionis2013multi, perdikaris2016multifidelity}, all face severe limitations in view of the non-Gaussian likelihoods and high-dimensional posterior distributions commonly encountered in realistic applications.

On the other hand, the recent explosive growth of machine learning research has put forth new effective ways of learning and manipulating complex high-dimensional probability distributions.  Inference tools like variational auto-encoders \cite{kingma2013auto} and generative adversarial networks \cite{goodfellow2014generative}, formulated on top of flexible building blocks such as feed-forward/convolutional/recurrent neural networks \cite{goodfellow2016deep}
have introduced highly scalable solutions, albeit for problems where not much prior information is assumed, but instead, large amounts of data can be obtained at relatively low cost (e.g., image recognition \cite{krizhevsky2012imagenet}, natural language processing \cite{lecun2015deep}).  In this work, we aim to leverage recent developments in machine learning to put forth a scalable framework for uncertainty propagation in physical systems for which 
the cost of data acquisition is high and training  data-sets  are  typically  small, but strong prior information exists by means of known governing laws expressed by partial differential equations. Specifically, we construct a class of probabilistic physics-informed neural networks that enables us to obtain a posterior characterization of the uncertainty associated with their predicted outputs. Moreover, we will develop a flexible variational inference framework that will allow us to train such models directly from noisy input/output data, and predict outcomes of non-linear dynamical systems that are partially observed with quantified uncertainty. This setting necessitates a departure from the classical deterministic realm of modeling and scientific computation, and, consequently, our main building blocks can no longer be crisp deterministic numbers and governing laws, but instead we must operate with {\em probabilistic models}.

This paper is structured as follows. In section \ref{sec:PINNs} we provide a brief overview of physics-informed neural networks in sync with the recent developments in \cite{raissi2017physics1,raissi2017physics2,raissi2018hidden,raissi2018deep}. In sections \ref{sec:PPINNs} and \ref{sec:ADVI} we provide an outline of the proposed probabilistic formulation and the proposed variational inference framework. Finally, in section \ref{sec:Results} we will demonstrate the effectiveness of our approach through a series of examples involving uncertainty propagation in non-linear conservation laws, and the discovery of constitutive laws for flow through porous media directly from noisy data.

\section{Methods}\label{sec:methods}

\subsection{Physics-informed neural networks}\label{sec:PINNs}
The recent works of Raissi {\em et. al.} \cite{raissi2017physics1,raissi2017physics2, raissi2018hidden,raissi2018deep} have  demonstrated how classical conservation laws and numerical discretization schemes can be used as structured prior information that can enhance the robustness and efficiency of modern machine learning algorithms, introducing a new class of data-driven solvers, as well as a {\em physics-informed machine learning} approach to model discovery. To this end, the authors have considered constructing deep neural networks that return predictions which are constrained by parametrized partial differential equations (PDE) of the form

\begin{equation}
\bm{u}_t + \mathcal{N}_{\bm{x}}\bm{u} = 0,
\end{equation}
where $\bm{u}(\bm{x},t)$ is represented by a deep neural network parametrized by a set of parameters $\theta$, i.e. $u(\bm{x},t) = f_{\theta}(\bm{x},t)$, $\bm{x}$ is a vector of space coordinates, $t$ is the time coordinate, and $\mathcal{N}_{\bm{x}}$ is a nonlinear differential operator. As neural networks are differentiable representations, this construction defines a so-called {\em physics informed neural network} that corresponds to the PDE residual, i.e. $r_{\theta}(\bm{x},t):= \frac{\partial}{\partial t} f_{\theta}(\bm{x},t)  + \mathcal{N}_{\bm{x}}f_{\theta}(\bm{x},t)$. This new network has the same parameters as the network representing $\bm{u}(\bm{x},t)$, albeit different activation functions due to the action of the differential operator \cite{raissi2017physics1, psichogios1992hybrid, lagaris1998artificial}. From an implementation perspective, this network can be readily obtained by leveraging recent progress in automatic differentiation \cite{baydin2015automatic, abadi2016tensorflow}.
The resulting training procedure allows us to recover the shared network parameters $\theta$ using a few scattered observations of $\bm{u}(\bm{x},t)$, namely $\{(\bm{x}_{i}, t_i), \bm{u}_i\}$, $i = 1,\dots,N_u$, along with a larger number of collocation points $\{(\bm{x}_{i}, t_i), \bm{r}_i = 0\}$, $i = 1,\dots,N_r$, that aim to penalize the PDE residual at a finite set of $N_r$ collocation nodes. This simple, yet remarkably effective regularization procedure allows us to introduce the PDE residual as a soft penalty constraint penalty in the likelihood function of the model \cite{raissi2017physics1,raissi2017physics2}, and the resulting optimization problem can be effectively solved using standard stochastic gradient descent without necessitating any elaborate constrained optimization techniques, simply by minimizing the composite loss function

\begin{equation}\label{eq:PINN_loss}
\mathcal{L}(\theta) = \frac{1}{N_u}\sum\limits_{i=1}^{N_u}\|f_{\theta}(\bm{x}_{i}, t_i) - \bm{u}_i\|^2  +   \frac{1}{N_r}\sum\limits_{i=1}^{N_r}\|r_{\theta}(\bm{x}_{i}, t_i) - \bm{r}_i\|^2, 
\end{equation}
where the required gradients $\frac{\partial \mathcal{L}}{\partial \theta}$ can be readily obtained using automatic differentiation \cite{baydin2015automatic}.
Finally, as the resulting predictions are encouraged to inherit any physical properties imposed by the PDE constraint (e.g., conservation, invariance, symmetries, etc.), this approach showcases how one can approximately encode physical and domain-specific constraints in modern machine learning algorithms and introduce a new form of regularization for learning from {\em small} data-sets.


\subsection{Probabilistic physics-informed neural networks}\label{sec:PPINNs}

Here we put forth a probabilistic formulation for propagating uncertainty through physics-informed neural networks using latent variable models of the form 

\begin{equation}
\label{eq:PSC}
p(\bm{u}|\bm{x},t,\bm{z}), \ \ \bm{z}\sim p(\bm{z}), \ \ \text{s.t} \ \  \bm{u}_t + \mathcal{N}_{\bm{x}}\bm{u} = 0
\end{equation}

This setting encapsulates a wide range of deterministic and stochastic problems, where $\bm{u}(\bm{x},t)$ is a potentially multi-variate field, and $\bm{z}$ is a collection of random latent variables. The ability to learn such a model from data is the cornerstone of probabilistic scientific computing and uncertainty quantification in physical systems. Knowledge of the conditional probability $p(\bm{u}|\bm{x},t,\bm{z})$ subject to domain knowledge constraints introduces a regularization mechanism that limits the space of admissible solutions to a manageable size (e.g., in fluid mechanics problems by discarding any non-realistic flow solutions that violate the conservation of mass principle), thus enables training of probabilistic deep learning algorithms in {\em small data} regimes.
Moreover, by providing a complete characterization of uncertainty, it enhances the robustness of our predictions, and provides a-posteriori error estimates for assessing model inadequacy. The latter, can also enable downstream tasks such as the formulation of adaptive data acquisition policies for active learning or Bayesian optimization \cite{shahriari2016taking} with domain knowledge constraints. Finally, thanks to the structure encoded by the PDE itself, the resulting latent variables $\bm{z}$  can potentially lead to the extraction of physically relevant and interpretable low-dimensional feature representations, which can  subsequently introduce new techniques for nonlinear model order reduction and coarse-graining of complex systems.


\subsection{Adversarial inference for joint distribution matching}\label{sec:ADVI}

Following the recent findings of \cite{li2018learning} we argue that matching the joint distribution of the generated data $p_{\theta}(\bm{x},t,\bm{u})$ with the joint distribution of the observed data $q(\bm{x},t,\bm{u})$ by 
 minimizing the reverse Kullback-Leibler divergence $\mathbb{KL}[p_{\theta}(\bm{x},t,\bm{u})||q(\bm{x},t,\bm{u})]$ is a promising approach to train the generative model presented in equation \ref{eq:PSC}. This also implies that the respective marginal and conditional distributions are also encouraged to match. The use of the reverse Kullback-Leibler divergence (in contrast to the maximum likelihood setup) is  motivated by examining the following decomposition

\begin{align}\label{eq:KL}
\mathbb{KL}[p_{\theta}(\bm{x},t,\bm{u})||q(\bm{x},t,\bm{u})] & = -h(p_{\theta}(\bm{x},t,\bm{u}))) - \mathbb{E}_{p_{\theta}(\bm{x},t,\bm{u})}[\log(q(\bm{x},t,\bm{u}))],
\end{align}
where $h(p_{\theta}(\bm{x},t,\bm{u}))$ denotes the entropy of the generative model. The second term can be further decomposed as

\begin{align}
\mathbb{E}_{p_{\theta}(\bm{x},t,\bm{u})}[\log(q(\bm{x},t,\bm{u}))]  = & \int_{\mathcal{S}_{p_{\theta}\cap \mathcal{S}_q}} \log(q(\bm{x},t,\bm{u})) p_{\theta}(\bm{x},t,\bm{u}) d\bm{x} dt d\bm{u} \ + \label{eq:decomposition} \\
& \int_{\mathcal{S}_{p_{\theta}\cap \mathcal{S}_q^{o}}} \log(q(\bm{x},t,\bm{u})) p_{\theta}(\bm{x},t,\bm{u}) d\bm{x} dt d\bm{u} \nonumber,
\end{align}
where $\mathcal{S}_{p_{\theta}}$ and $\mathcal{S}_q$ denote the support of the distributions $p_{\theta}(\bm{x},t,\bm{u})$ and $q(\bm{x},t,\bm{u})$, respectively, while $\mathcal{S}_q^{o}$ denotes the complement of $\mathcal{S}_q$. Notice that by minimizing the Kullback-Leibler divergence in equation \ref{eq:KL} we introduce a mechanism that is trying to balance the effect of two competing objectives. Specifically, maximization of the entropy term $h(p_{\theta}(\bm{x},t,\bm{u})))$ encourages $p_{\theta}(\bm{x},t,\bm{u})$ to spread over its support set as wide, while the second integral term in equation \ref{eq:decomposition} introduces a strong (negative) penalty when the support of $p_{\theta}(\bm{x},t,\bm{u})$ and $q(\bm{x},t,\bm{u})$ do not overlap. Hence, the support of $p_{\theta}(\bm{x},t,\bm{u})$ is encouraged to spread only up to the point that $\mathcal{S}_{p_{\theta}}\cap \mathcal{S}_{q^{o}} = \emptyset$, implying that $\mathcal{S}_{p_{\theta}}\subseteq \mathcal{S}_{q^{o}}$. When $\mathcal{S}_{p_{\theta}}\subset \mathcal{S}_{q^{o}}$ the pathological issue of ``mode-collapse" (commonly encountered in the training of generative adversarial networks \cite{goodfellow2014generative}) is manifested \cite{salimans2016improved}. 
This issue is present if one seeks to directly minimize the reverse Kullback-Leibler objective in equation \ref{eq:KL} as this provides no control on the relative importance of the two
terms. As discussed in \cite{li2018learning},  we may rather minimize $-\lambda h(p_{\theta}(\bm{x},t,\bm{u}))) - \mathbb{E}_{p_{\theta}(\bm{x},t,\bm{u})}[\log(q(\bm{x},t,\bm{u}))]$, with $\lambda \ge 1$ to allow for control of how much
emphasis is placed on mitigating mode collapse. It is then clear that the entropic regularization introduced by $h(p_{\theta}(\bm{x},t,\bm{u})))$ provides an effective mechanism for controlling and mitigating the effect of mode collapse, and, therefore, potentially enhancing the robustness adversarial inference procedures for learning $p_{\theta}(\bm{x},t,\bm{u})$. 

Minimization of equation \ref{eq:KL} with respect to the generative model parameters $\theta$ presents two fundamental difficulties. First, the evaluation of both distributions $p_{\theta}(\bm{x},t,\bm{u})$ and $q(\bm{x},t,\bm{u})$ typically involves intractable integrals in high dimensions, and we may only have samples drawn from the two distributions, not their explicit analytical forms. Second, the differential entropy term $h(p_{\theta}(\bm{x},t,\bm{u})))$ is intractable as $p_{\theta}(\bm{x},t,\bm{u}))$ is not known a-priori. In the next sections we revisit the unsupervised formulation put forth in \cite{li2018learning} and derive a tractable inference procedure for learning $p_{\theta}(\bm{x},t,\bm{u}))$ from scattered observation pairs of $\bm{u}(\bm{x},t)$, namely $\{(\bm{x}_{i}, t_i), \bm{u}_i\}$, $i = 1,\dots,N_u$.

\subsubsection{Density ratio estimation by probabilistic classification}\label{sec:density_ratio}

By definition, the computation of the Kullback-Leibler divergence in equation \ref{eq:KL} involves computing an expectation over a log-density ratio, i.e. 
$$\mathbb{KL}[p_{\theta}(\bm{x},t,\bm{u})||q(\bm{x},t,\bm{u})] := \mathbb{E}_{p_{\theta}(\bm{x},t,\bm{u})}\left[\log\left(\frac{p_{\theta}(\bm{x},t,\bm{u})}{q(\bm{x},t,\bm{u})}\right)\right].$$ 
In general, given samples from two distributions, we can approximate their density ratio by constructing a binary classifier that distinguishes between samples from the two distributions. To this end, we assume that  $N$ data points are drawn from $p_{\theta}(\bm{x},t,\bm{u})$ and are assigned a label $y=+1$. Similarly, we assume that $N$ samples are drawn from  $q(\bm{x},t,\bm{u})$ and assigned label $y=-1$. Consequently, we can write these probabilities in a conditional form, namely
$$p_{\theta}(\bm{x},t,\bm{u}) = \rho(\bm{x},t,\bm{u}|y=+1), \ \ q(\bm{x},t,\bm{u}) = \rho(\bm{x},t,\bm{u}|y=-1),$$
where $\rho(\bm{x},t,\bm{u}|y=+1)$ and $\rho(\bm{x},t,\bm{u}|y=-1)$ are the class probabilities predicted by a binary classifier $T(\bm{x},t,\bm{u})$.
Using Bayes rule, it is then straightforward to show that the density ratio of $p_{\theta}(\bm{x},t,\bm{u})$ and $q(\bm{x},t,\bm{u})$ can be computed as

\begin{align}
    \frac{p_{\theta}(\bm{x},t,\bm{u})}{ q(\bm{x},t,\bm{u})} & = \frac{\rho(\bm{x},t,\bm{u}|y=+1)}{\rho(\bm{x},t,\bm{u}|y=-1)} \nonumber\\
    & = \frac{\rho(y=+1|\bm{x},t,\bm{u})\rho(\bm{x},t,\bm{u})}{\rho(y=+1)} \bigg/ \frac{\rho(y=-1|\bm{x},t,\bm{u})\rho(\bm{x},t,\bm{u})}{\rho(y=-1)} \nonumber\\
    & = \frac{\rho(y=+1|\bm{x},t,\bm{u})}{\rho(y=-1|\bm{x},t,\bm{u})} = \frac{\rho(y=+1|\bm{x},t,\bm{u})}{1 - \rho(y=+1|\bm{x},t,\bm{u})} \nonumber\\
    & = \frac{T(\bm{x},t,\bm{u})}{1-T(\bm{x},t,\bm{u})}.
\end{align}
This simple procedure suggests that we can harness the power of deep neural network classifiers to obtain accurate estimates of the reverse Kullback-Leibler divergence in equation \ref{eq:KL} directly from data and without the need to assume any specific parametrization for the generative model distribution $p_{\theta}(\bm{x},t,\bm{u})$.

\subsubsection{Entropic regularization bound}\label{sec:entropy_bound}
Here we follow the derivation of Li {\em et. al} \cite{li2018learning} to construct a computable lower bound for the entropy $h(p_{\theta}(\bm{x},t,\bm{u}))$. To this end, we start by considering random variables $(\bm{x}, t, \bm{u}, \bm{z})$ under the joint distribution 
$$p_{\theta}(\bm{x}, t, \bm{u}, \bm{z}) = p_{\theta}(\bm{u}, \bm{x}, t|\bm{z}) p(\bm{z}) = p_{\theta}(\bm{u}|\bm{x}, t, \bm{z}) p(\bm{x}, t, \bm{z}),$$ 
where $p_{\theta}(\bm{u}|\bm{x}, t, \bm{z}) = \delta(\bm{u}-f_{\theta}(\bm{x}, t, \bm{z}))$, and $\delta(\cdot)$ is the Dirac delta function. The mutual information between $(\bm{x}, t, \bm{u})$ and $\bm{z}$ satisfies the information theoretic identity
$$ I(\bm{x}, t, \bm{u}; \bm{z}) = h(\bm{x}, t, \bm{u})-h(\bm{x}, t, \bm{u}|\bm{z}) =
h(\bm{z}) - h(\bm{z}|\bm{x}, t, \bm{u}),$$
where $h(\bm{x}, t, \bm{u})$, $h(\bm{z})$ are the marginal entropies and $h(\bm{x}, t, \bm{u}|\bm{z})$, $h(\bm{z}|\bm{x}, t, \bm{u})$ are the conditional entropies \cite{akaike1998information}. 
Since in our setup $\bm{x}$ and $t$ are deterministic variables independent of $\bm{z}$, and samples of $p_{\theta}(\bm{u}|\bm{x}, t, \bm{z})$ are generated by a deterministic function $f_{\theta}(\bm{x}, t, \bm{z})$, it follows that $h(\bm{x}, t, \bm{u}|\bm{z}) = 0$. 
We therefore have
\begin{equation}\label{eq:info_identity}
h(\bm{x}, t, \bm{u}) = h(\bm{z}) - h(\bm{z}|\bm{x}, t, \bm{u}),
\end{equation}
where $h(\bm{z}) := -\int \log p(\bm{z}) p(\bm{z}) d\bm{z}$ does not depend on the generative model parameters $\theta$. 

Now consider a general variational distribution $q_{\phi}(\bm{z}|\bm{x}, t, \bm{u})$ parametrized by a set of parameters $\phi$. Then,

\begin{align}
   h(\bm{z}|\bm{x}, t, \bm{u})  = & -\mathbb{E}_{p_{\theta}(\bm{x}, t, \bm{u}, \bm{z})}[\log(p_{\theta}(\bm{z}|\bm{x}, t, \bm{u}))] \nonumber\\
    = & -\mathbb{E}_{p_{\theta}(\bm{x}, t, \bm{u}, \bm{z})}[\log(q_{\phi}(\bm{z}|\bm{x}, t, \bm{u}))] \nonumber  \\ 
    & -\mathbb{E}_{p_{\theta}(\bm{x}, t, \bm{u})}[\mathbb{KL}[p_{\theta}(\bm{z}|\bm{x}, t, \bm{u})||q_{\phi}(\bm{z}|\bm{x}, t, \bm{u})]] \nonumber \\
    \le & -\mathbb{E}_{p_{\theta}(\bm{x}, t, \bm{u}, \bm{z})}[\log(q_{\phi}(\bm{z}|\bm{x}, t, \bm{u}))]. \label{eq:entropy_bound}
\end{align}
Viewing $\bm{z}$ as a set of latent variables, then $q_{\phi}(\bm{z}|\bm{x}, t, \bm{u})$ is a variational approximation to the true intractable posterior over the latent variables $p_{\theta}(\bm{z}|\bm{x}, t, \bm{u})$. Therefore, 
if $q_{\phi}(\bm{z}|\bm{x}, t, \bm{u})$  is introduced as an auxiliary inference model associated with the generative model $p_{\theta}(\bm{x}, t, \bm{u})$, for which $\bm{u} = f_{\theta}(\bm{x}, t,\bm{z})$ and $\bm{z}\sim p(\bm{z})$, then we can use equations \ref{eq:info_identity} and \ref{eq:entropy_bound} to bound the entropy term in equation \ref{eq:KL} as
\begin{equation}
    h(p_{\theta}(\bm{x},t,\bm{u})) \ge h(p(\bm{z})) + \mathbb{E}_{p_{\theta}(\bm{x}, t, \bm{u}, \bm{z})}[\log(q_{\phi}(\bm{z}|\bm{x}, t, \bm{u}))].
\end{equation}
Note that the inference model $q_{\phi}(\bm{z}|\bm{x}, t, \bm{u})$ plays the role of a variational approximation to the true posterior over the latent variables, and appears naturally using information theoretic arguments in the derivation of the lower bound.

\subsubsection{Adversarial training objective}

By leveraging the density ratio estimation procedure described in section \ref{sec:density_ratio} and the entropy bound derived in section \ref{sec:entropy_bound}, we can derive the following loss functions for minimizing the reverse Kullback-Leibler divergence with entropy regularization

\begin{align}
	\mathcal{L}_{\mathcal{D}}(\psi) = & \  \mathbb{E}_{q(\bm{x},t)p(\bm{z})}[\log\sigma(T_{\psi}(\bm{x},t,f_{\theta}(\bm{x},t,\bm{z})))] + \nonumber \\ & \ \mathbb{E}_{q(\bm{x},t,\bm{u})}[\log(1-\sigma(T_{\psi}(\bm{x},t,\bm{u})))] \label{eq:discriminator_loss}\\
	\mathcal{L}_{\mathcal{G}}(\theta, \phi) = & \ \mathbb{E}_{q(\bm{x},t)p(\bm{z})}[T_{\psi}(\bm{x}, t, f_{\theta}(\bm{x},t,\bm{z}))+ (1-\lambda)\log(q_{\phi}(\bm{z}|\bm{x},t,f_{\theta}(\bm{x},t,\bm{z})))] \label{eq:generator_loss},
\end{align}
where $\sigma(x)=1/(1+e^{-x})$ is the logistic sigmoid function. Notice how the binary cross-entropy objective of equation \ref{eq:discriminator_loss} aims to progressively improve the ability of the classifier $T_{\psi}(\bm{x},t,\bm{u})$ to discriminate between ``fake" samples $(\bm{x},t,f_{\theta}(\bm{x},t,\bm{z}))$ obtained from the generative model $p_{\theta}(\bm{x},t,\bm{u})$ and ``true" samples $(\bm{x},t,\bm{u})$ originating from the observed data distribution $q(\bm{x},t,\bm{u})$. Simultaneously, the objective of equation \ref{eq:generator_loss} aims at improving the ability of the generator $f_{\theta}(\bm{x},t,\bm{u})$ to generate increasingly more realistic samples that can ``fool" the discriminator $T_{\psi}(\bm{x},t,\bm{u})$. Moreover, the encoder $q_{\phi}(\bm{z}|\bm{x},t,f_{\theta}(\bm{x},t,\bm{z}))$ not only serves as an entropic regularization term than allows us to stabilize model training and mitigate the pathology of mode collapse, but also provides a variational approximation to true posterior over the latent variables. The way it naturally appears in the objective of equation \ref{eq:generator_loss} also encourages the cycle-consistency of the latent variables $\bm{z}$; a process that is known to result in disentangled and interpretable low-dimensional representations of the observed data \cite{friedman2001elements}, which could be subsequently used as good features for nonlinear model order reduction.

In theory, the optimal set of parameters $\{\theta^{\ast}, \phi^{\ast}, \psi^{\ast}\}$ correspond to the Nash equilibrium of the two player game defined by the loss functions in equations \ref{eq:discriminator_loss},\ref{eq:generator_loss}, for which one can show that the exact model distribution and the exact posterior over the latent variables can be recovered \cite{goodfellow2014generative, pu2017symmetric}. In practice, although there is no guarantee that this optimal solution can be attained, the generative model can be trained by alternating between optimizing the two objectives in equations \ref{eq:discriminator_loss},\ref{eq:generator_loss} using stochastic gradient descent as

\begin{align}
    & \mathop{\max}_{\psi} \ \mathcal{L}_{\mathcal{D}}(\psi) \label{eq:discriminator}\\
	& \mathop{\min}_{\theta, \phi} \ \mathcal{L}_{\mathcal{G}}(\theta, \phi) \label{eq:generator}.
\end{align}

\subsubsection{Adversarial training with physics-informed constrains}

In order to learn the physics-informed probabilistic model of equation \ref{eq:PSC} from data we can extend the adversarial  inference framework presented above by appropriately penalizing the loss function of the generator (see equation \ref{eq:generator_loss}). The available data correspond to scattered observation pairs $\{(\bm{x}_{i}, t_i), \bm{u}_i\}$, $i = 1,\dots,N_u$, originating from known initial or boundary conditions, or any other (potentially noisy) measurements of $\bm{u}(\bm{x},t)$. In analogy to the deterministic setting put for in \cite{raissi2017physics1} and summarized in section \ref{sec:PINNs}, by defining $\bm{r}_{\theta}(\bm{x},t,\bm{z}):= \frac{\partial}{\partial t} f_{\theta}(\bm{x},t,\bm{z})  + \mathcal{N}_{\bm{x}}f_{\theta}(\bm{x},t,\bm{z})$ we essentially introduce a new conditional probability model $p_{\theta}(\bm{r}|\bm{x},t,\bm{z})$ that shares the same parameters as $p_{\theta}(\bm{u}|\bm{x},t,\bm{z})$, albeit the underlying neural network that serves as its approximation has different activation functions. However, since we would like to encourage every sample $\bm{u} = f_{\theta}(\bm{x},t,\bm{z})$ produced by the generator to satisfy the PDE constraint, we can simply treat the residual as a deterministic variable, i.e,  $\bm{r}_{\theta}(\bm{x},t,\bm{z}) = \bm{r}_{\theta}(\bm{x},t)$, and enforce the constraint at a finite set of collocation points $N_r$ by simply minimizing the mean square loss
\begin{equation}
\label{eq:PI_loss}
	\mathcal{L}_{PDE}(\theta) = \frac{1}{N_r}\sum_{i=1}^{N_r}\|\bm{r}_{\theta}(\bm{x}_i, t_i) - \bm{r}_i\|^2.
\end{equation}
Then, the resulting adversarial game for training the physics-informed model of equation \ref{eq:PSC} takes the form

\begin{equation}
    \label{eq:objective}
    \begin{aligned}
    	& \mathop{\max}_{\psi} \ \mathcal{L}_{\mathcal{D}}(\psi)\\
    	& \mathop{\min}_{\theta, \phi} \ \mathcal{L}_{\mathcal{G}}(\theta, \phi) + \beta \mathcal{L}_{PDE}(\theta),
    \end{aligned}
\end{equation}
where positive values of $\beta$ can be selected to place more emphasis on penalizing the PDE residual. For $\beta>0$, the residual loss $\mathcal{L}_{PDE}(\theta)$ acts as a regularization term that approximately enforces the given physical constraint, and, therefore, encourages the generator $p_{\theta}(\bm{u}|\bm{x},t,\bm{z})$ to produce samples that satisfy the underlying partial differential equation. Also note that this structured approach also encourages the encoder $q_{\phi}(\bm{z}|\bm{x},t,f_{\theta}(\bm{x},t,\bm{z}))$ to learn a set of spatio-temporal latent variables $\bm{z}$ that are relevant to the underlying physics, possibly opening a new directions for probabilistic model order reduction of complex systems.

\subsubsection{Predictive distribution}

Once the model is trained we can construct a probabilistic ensemble for the solution $\bm{u}(\bm{u}|\bm{x}, t, \bm{z})$ by sampling latent variables from the prior $p(\bm{z})$ and passing them through the generator to yield samples $\bm{u} = f_{\theta}(\bm{x}, t, \bm{z})$ that are distributed according to the predictive model distribution $p_{\theta}(\bm{u}|\bm{x}, t, \bm{z})$. Note that although the explicit form of this  distribution is not known, we can efficiently compute any of its moments via Monte Carlo sampling. The cost of this prediction step is negligible compared to the cost of training the model, as it only involves a single forward pass through the generator function $f_{\theta}(\bm{x}, t, \bm{z})$. Typically, we compute the mean and variance of the predictive distribution at a new test point $(\bm{x}^{\ast}, t^{\ast})$ as 

\begin{align}
    \mu_{\bm{u}}(\bm{x}^{\ast}, t^{\ast}) & = \mathbb{E}_{p_{\theta}}[\bm{u}|\bm{x}^{\ast}, t^{\ast}, \bm{z}] \approx \frac{1}{N_s}\sum\limits_{i=1}^{N_s} f_{\theta}(\bm{x}^{\ast}, t^{\ast}, \bm{z}_i), \label{eq:predictive_mean} \\
    \sigma^{2}_{\bm{u}}(\bm{x}^{\ast}, t^{\ast}) & = \mathbb{V}\text{ar}_{p_{\theta}}[\bm{u}|\bm{x}^{\ast}, t^{\ast}, \bm{z}] \approx \frac{1}{N_s}\sum\limits_{i=1}^{N_s} [f_{\theta}(\bm{x}^{\ast}, t^{\ast}, \bm{z}_i) - \mu_{\bm{u}}(\bm{x}^{\ast}, t^{\ast})]^2, \label{eq:predictive_variance}
\end{align}
where $\bm{z}_i \sim p(\bm{z})$, $i = 1,\dots,N_s$, and $N_s$ corresponds to the total number of Monte Carlo samples.

\subsubsection{Advantages and caveats of adversarial learning}

Since their recent introduction \cite{goodfellow2014generative, makhzani2015adversarial, dumoulin2016adversarially, mescheder2017adversarial}, adversarial learning  techniques have provided great flexibility for performing probabilistic computations with arbitrarily complex implicit distributions. Essentially, they have lifted  the over-simplified approximations typically used in variational inference (Gaussian approximations, exponential families, etc.) \cite{blei2017variational}, yielding very general and flexible schemes for statistical inference. However, this flexibility comes at a price, as such methods in practice require very careful tuning in order to achieve stable and accurate performance. To this end, recall the training objective defined in equation \ref{eq:objective} that introduces an adversarial game between the generator and discriminator networks \cite{goodfellow2014generative}. In practice, this mini-max optimization problem is solved by alternating stochastic gradient updates between the two competing objectives, and it is highly sensitive on the capacity of the neural networks modeling the generator and discriminator, as well as the relative frequency with which the parameters of each network are updated within each iteration of stochastic gradient descent. To this end, we provide a series of empirical observations and lessons we learned throughout this study that can enhance the robustness and stability of this training procedure:

\begin{itemize}
    \item Changing the relative number of stochastic gradient updates for the generator $K_g$ and the discriminator $K_d$ is equivalent to changing their neural network architecture. For example, we can reduce the capacity of discriminator by either performing more stochastic gradient updates for the generator, or remove one layer in the neural network architecture of the discriminator. 
    \item Given enough collocation points $N_r$ for penalizing the PDE residual, we can obtain robust uncertainty estimates together with precise predictions simply by tuning the capacity of discriminator and generator networks.
    \item Typically, by fixing the generator, we expect the discriminator to have some capacity so that the model training dynamics remain stable. But, we do not want the discriminator to be very powerful as in that case there will by very little information from the discriminator that can help the generator to improve towards producing more realistic samples (this a common characteristic of adversarial inference procedures \cite{goodfellow2014generative}). 
    \item  For cases with a small number of training data we should reduce the capacity of the discriminator.  This can be achieved by either changing the relative frequency of stochastic gradient updates for the generator and discriminator, or by reducing the capacity of the discriminator neural network architecture.
\end{itemize}



\section{Results}
\label{sec:Results}

In all examples we have trained the models for 30,000 stochastic gradient descent steps using the Adam optimizer \cite{kingma2014adam} with a learning rate of $10^{-4}$, while fixing a one-to-five ratio for the discriminator versus generator updates. Moreover, we have fixed the entropic regularization and the residual penalty parameters to $\lambda = 1.5$ and $\beta = 1.0$, respectively. The proposed algorithms were implemented in Tensorflow v1.10 \cite{abadi2016tensorflow}, and computations were performed in single precision arithmetic on a single NVIDIA Tesla P100 GPU card. All data and code accompanying this manuscript will be made available at \url{https://github.com/PredictiveIntelligenceLab/UQPINNs}.

\subsection{A pedagogical example}

Let us illustrate the basic capabilities of the proposed methods through a simple example corresponding to the following nonlinear second-order ordinary differential equation
\begin{equation}\label{eq:ODE}
\begin{aligned}
	&u_{xx} - u^2u_x = f(x), \quad\quad x\in[-1, 1], \\
	&f(x) = -\pi^2\sin(\pi x) - \pi\cos(\pi x)\sin^2(\pi x),\\
\end{aligned}
\end{equation}
subject to random boundary conditions $u(-1), u(1)\sim \mathcal{N}(\bm{0}, \sigma_n^2\bm{I})$. 
For this simple example, the deterministic solution corresponding to $\sigma_n^2 = 0$ can be readily obtained as $u(x) = \sin(\pi x)$. Given $N_u$ observations of $u(x)$ corresponding to different realizations of the random boundary conditions our goal is to obtain a probabilistic representation of the solution $p_{\theta}(u|x,\bm{z})$ by training a physics-informed generative model of the form $u = f_{\theta}(x,\bm{z})$, $\bm{z}\sim p(\bm{z})$ that is constrained by equation \ref{eq:ODE}. To this end, we introduce three deterministic mappings parametrized by deep neural networks, namely $f_{\theta}(x, \bm{z})$, $q_{\phi}(x, u)$, and $T_{\psi}(x, u)$ corresponding to the generator, encoder, and discriminator functions introduced in section \ref{sec:ADVI}. By construction, we also obtain a physics-informed neural network $r_{\theta}(x)$ corresponding to the deterministic residual of equation \ref{eq:ODE} that will be used to approximately enforce the differential equation constraint at a set of $N_r = 100$ randomly distributed collocation points $x\in[-1, 1]$. All neural networks were chosen to have two hidden layers with 50 neurons in each layer, and a hyperbolic tangent activation function. Moreover, the dimensionality of the latent variables was set to one, i.e. $\bm{z}=z$, and we have assumed an isotropic standard normal prior, namely $p(z)\sim \mathcal{N}(0,1)$. As the training data for $u(x)$ reflects the uncertainty in the boundary conditions, the role of the latent variables $z$ is to enable the propagation of  this uncertainty into the predicted solution obtained through the generative model $p_{\theta}(u|x,z)$.

\begin{figure}
\centering
\includegraphics[width=\textwidth]{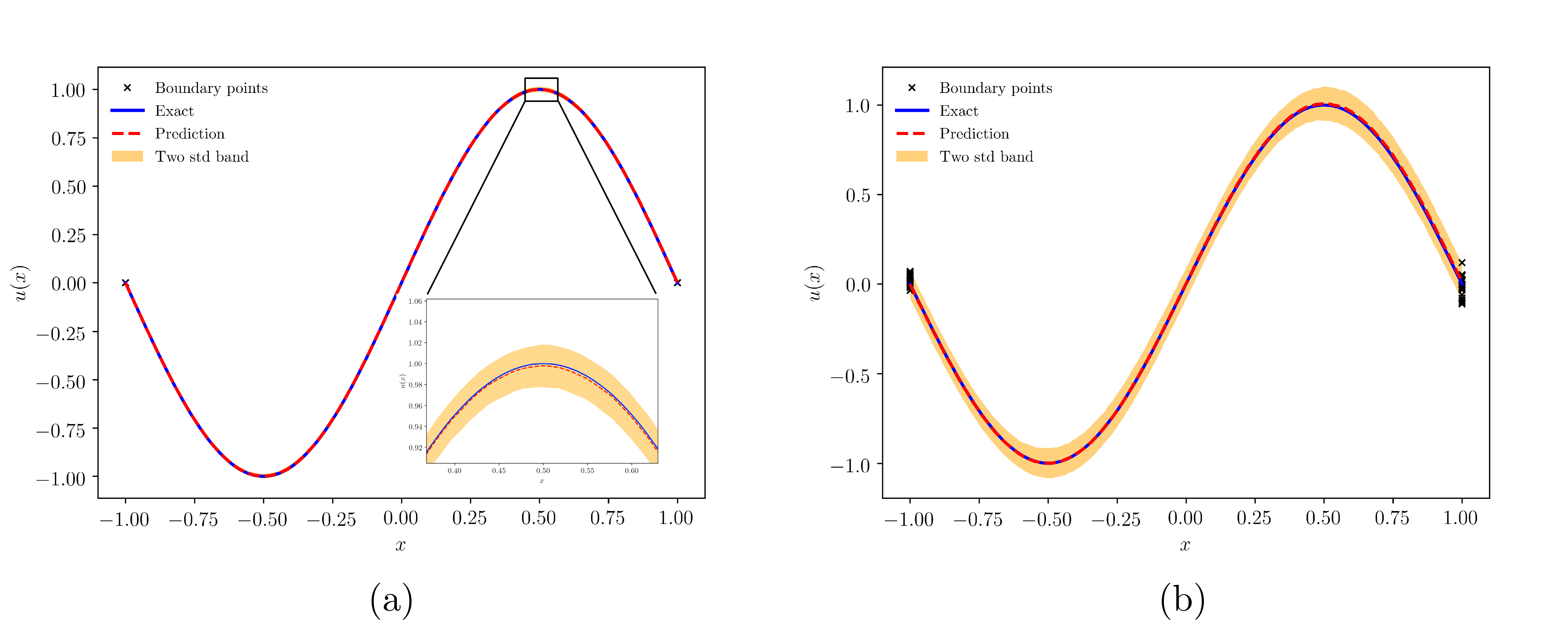}
\caption{{\em A pedagogical example:} (a) Mean and two standard deviations of $p_{\theta}(u|x,z)$ against the exact solution for deterministic boundary data. (b) Mean and two standard deviations of $p_{\theta}(u|x,z)$ against the reference Monte Carlo solution for random boundary data corresponding to $5\%$ Gaussian uncorrelated noise.}
\label{fig:ODE_solution}
\end{figure}

\begin{figure}
\centering
\includegraphics[width=\textwidth]{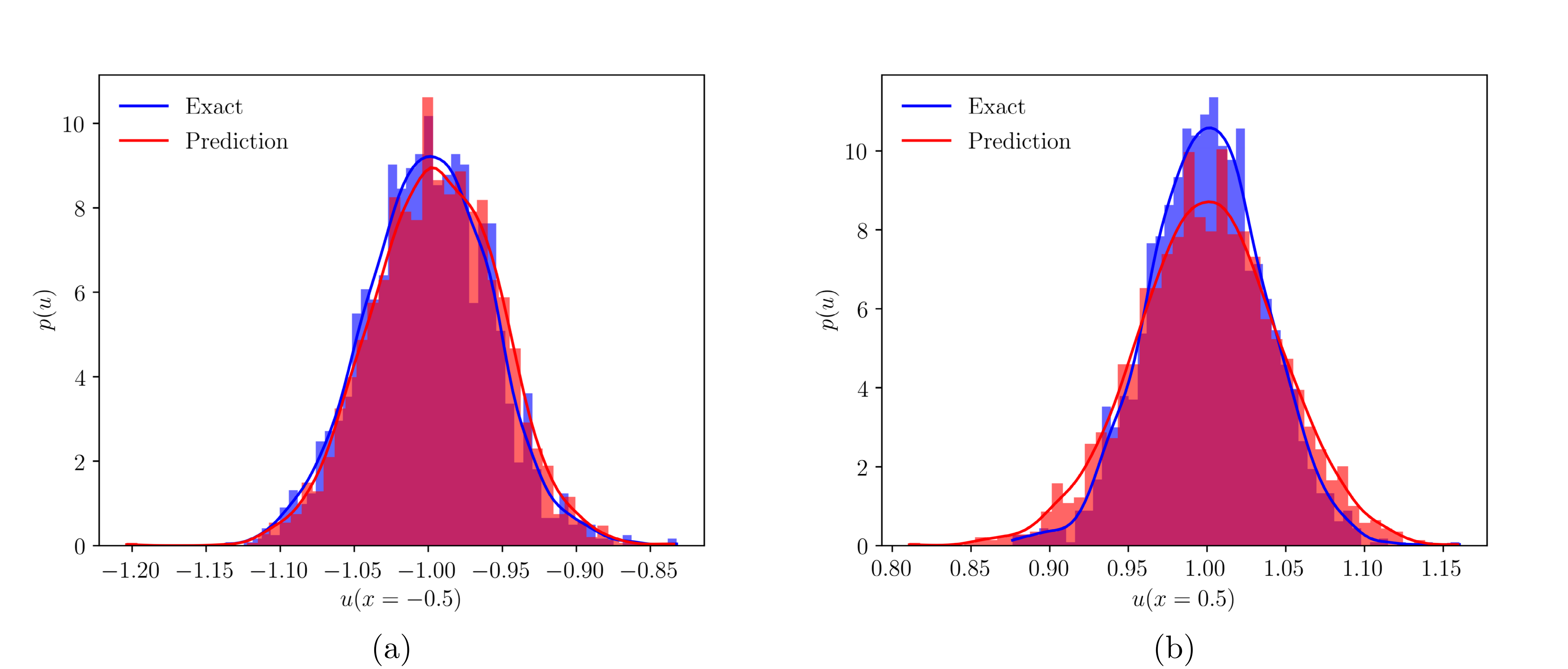}
\caption{{\em A pedagogical example:} Predicted marginal densities against the reference Monte Carlo solution. (a) $p_{\theta}(u|x=-0.5,z)$. (b) $p_{\theta}(u|x=+0.5,z)$.}
\label{fig:ODE_density}
\end{figure}

Here we have considered two cases corresponding to deterministic and random boundary conditions, namely (i) $\sigma_n^2=0$ (i.e., noise-free data), and (ii) $\sigma_n^2= 0.05$ (i.e., $\%5$ Gaussian uncorrelated noise). In all cases, the training data consists of $N_u=\textcolor{black}{20}$ realizations for each boundary point, $u(-1), u(1)$, and a total of $N_r=100$ collocation points for enforcing the residual of equation \ref{eq:ODE}. Our probabilistic predictions for this example are summarized in figures \ref{fig:ODE_solution} and \ref{fig:ODE_density}. Specifically, figure \ref{fig:ODE_solution}(a) shows the generative model predictive mean and two standard deviations, plotted against the exact solution of this problem. Note that this case corresponds to deterministic training data for the boundary conditions, hence the exact solution is deterministic, and the prediction error here is measured as $\mathcal{E}_{\mathbb{L}_2} = 1.36\cdot10^{-3}$ in the relative $\mathbb{L}_2$ norm

\begin{equation}
    \mathcal{E}_{\mathbb{L}_2} := \frac{\sqrt{\sum\limits_{i=1}^{N^{\ast}} [\mu_{\bm{u}}(x_i^{\ast}) - u(x_i^{\ast})]^2}}{\sqrt{\sum\limits_{i=1}^{N^{\ast}} [u(x_i^{\ast})]^2}},
\end{equation}
where $N^{\ast} = \textcolor{black}{200}$ denotes the total number of equidistant test points $x^{\ast}$ in the interval $[-1, 1]$. Moreover, the variance shown in the inset of figure \ref{fig:ODE_solution}(a) serves as an {\it a-posteriori} error estimate  that quantifies the uncertainty associated with the generative model predictions. Figure \ref{fig:ODE_solution}(b) shows the resulting prediction and uncertainty estimates corresponding to random boundary conditions, compared against a reference mean solution obtained numerically using a spectral method with \textcolor{black}{2,000} Monte Carlo samples. In this case, the predictive uncertainty of the generative model reflects the aggregate {\em total} uncertainty due to both randomness in the boundary conditions and the inherent epistemic uncertainty in the neural network approximation. As the generative model can return a complete statistical characterization of the solution by means of its conditional probability density $p_{\theta}(u|x,z)$, in figure \ref{fig:ODE_density} we provide a visual comparison of the one-dimensional marginals between our predictions and the reference Monte Carlo solution corresponding to the spatial locations $x = -0.5$ and $x = 0.5$.  

Albeit simple, this example aims to demonstrate the basic capabilities of the proposed methodology in propagating uncertainty through non-linear partial differential equations. In contrast to previous approaches to inferring solutions of partial differential equations from data \cite{cockayne2016probabilistic,raissi2017inferring,raissi2018numerical,cockayne2017bayesian}, the proposed methodology does not rely on Gaussian assumptions, and it can directly tackle nonlinear problems without any need for linearization.

\subsection{Burgers equation}\label{sec:burgers}
In this example we aim to provide a comprehensive systematic study to quantify the robustness of the proposed methods with respect to different parameter choices. We will do so through the lens of a more challenging canonical problem involving the non-linear time-dependent Burgers equation in one spatial dimension:
\begin{equation}
\label{eq:Burgers}
\begin{aligned}
	&u_t + u u_x - \nu u_{xx} = 0, \quad\quad x\in[-1, 1], t\in[0,1],\\
	&u(0,x) = -\sin(\pi x),\\
	&u(t,-1) = u(t,1) = 0,\\
\end{aligned}
\end{equation}
where the viscosity parameter is chosen as $\nu = (0.01/\pi)$ in order to generate a strongly nonlinear response that leads to the development of shock discontinuities in finite time. This is one of the few nonlinear partial differential equations that admits an exact solution through the Cole-Hopf transformation \cite{hopf1950partial}; a solution that will be subsequently used to test the validity of our predictions. 

Here we represent the unknown solution $u(x,t)$ using a physics-informed generative model of the form $u = f_{\theta}(x, \textcolor{black}{t}, \bm{z})$, and we will introduce parametric functions corresponding to a generator $f_{\theta}(x, t, \bm{z})$, an encoder $q_{\phi}(x, t, u)$, and a discriminator $T_{\psi}(x, \textcolor{black}{t}, u)$ all constructed using deep feed-forward neural networks. The baseline architectures for the generator and the encoder have 4 hidden layers with 50 neurons per layers, while the discriminator network has 3 hidden layers and 50 neurons per layer. The activation function in all cases is chosen to be a hyperbolic tangent non-linearity. The prior over the latent variables $p(\bm{z})$ is chosen again to be a one-dimensional isotropic Gaussian distribution, i.e. $\bm{z} = z$, $z \sim \mathcal{N}(0,1)$.

\begin{figure}
\centering
\includegraphics[width=\textwidth]{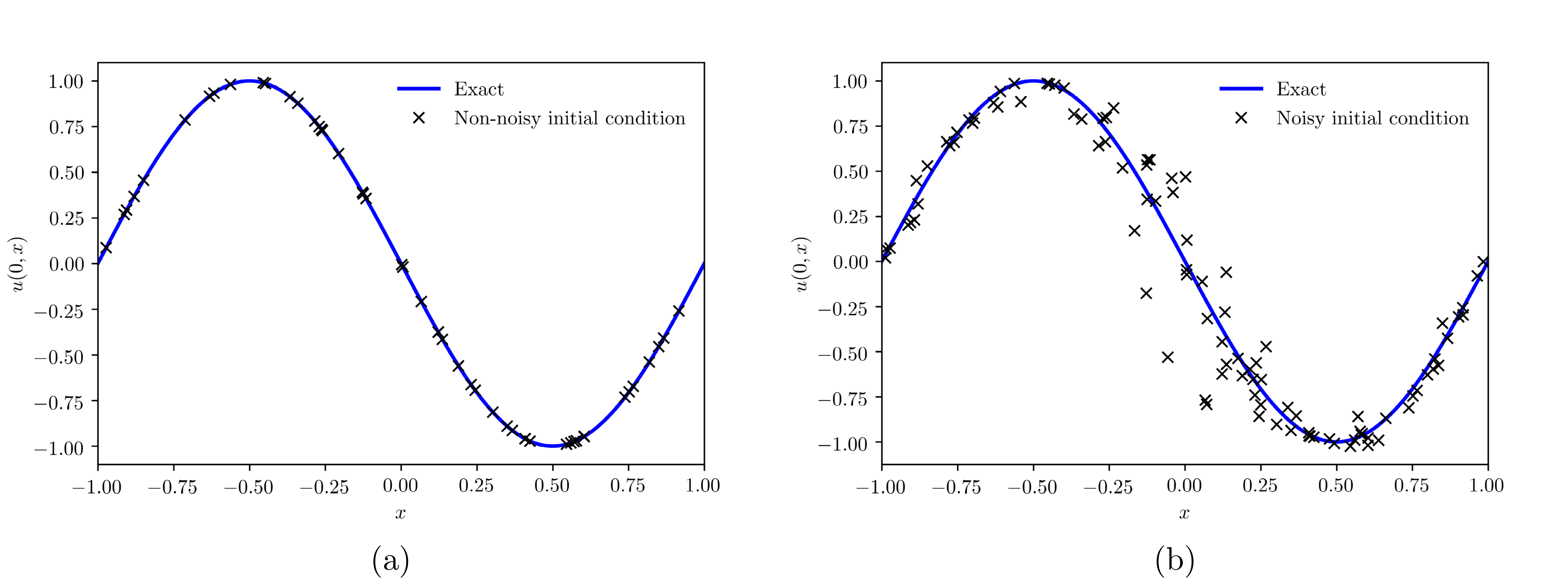}
\caption{{\em Burgers equation:} (a) Exact initial condition and noise-free training data (50 points). (b) Training data corresponding to a single realization of the non-additive noise corruption process (100 points, generated by equation \ref{eq:noisy_data}).}
\label{fig:Burgers_initial}
\end{figure}

\begin{figure}
\centering
\includegraphics[width=\textwidth]{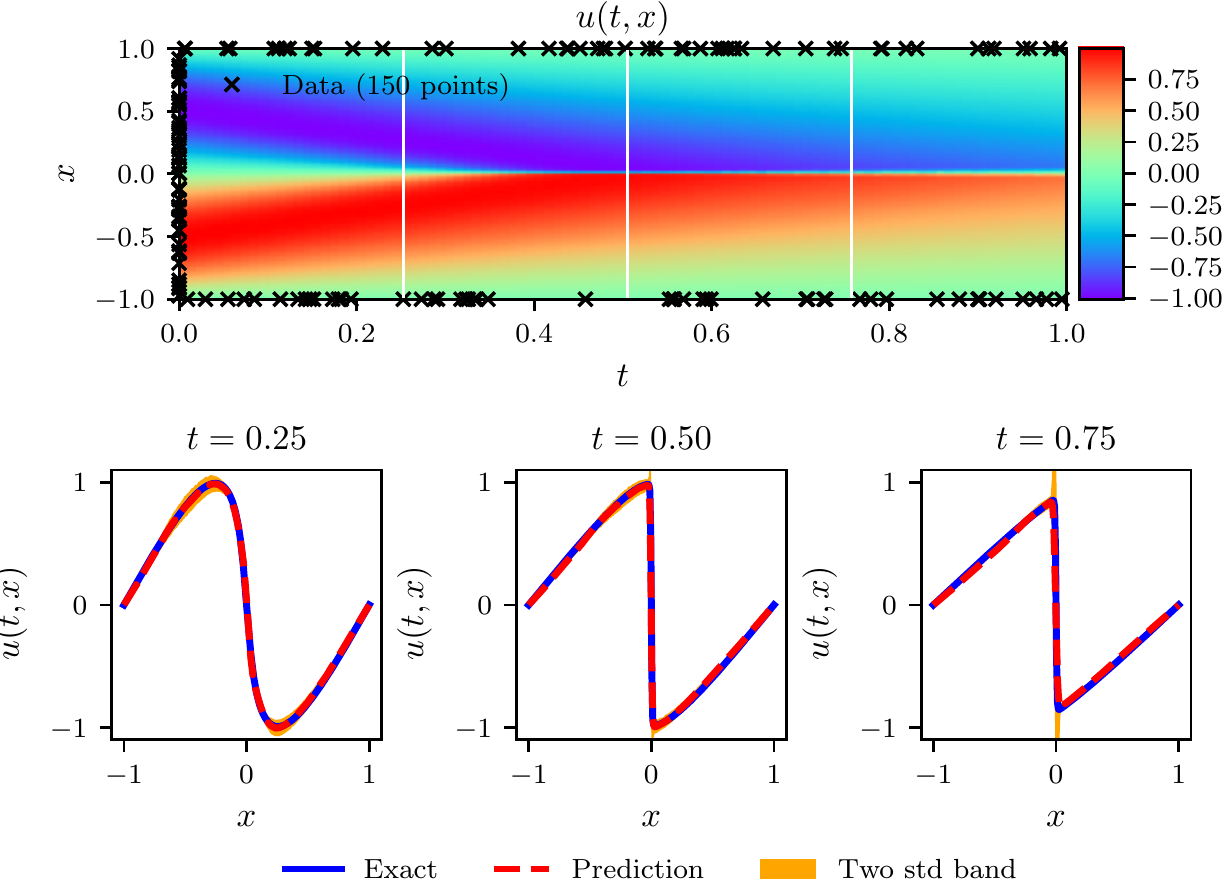}
\includegraphics[width=\textwidth]{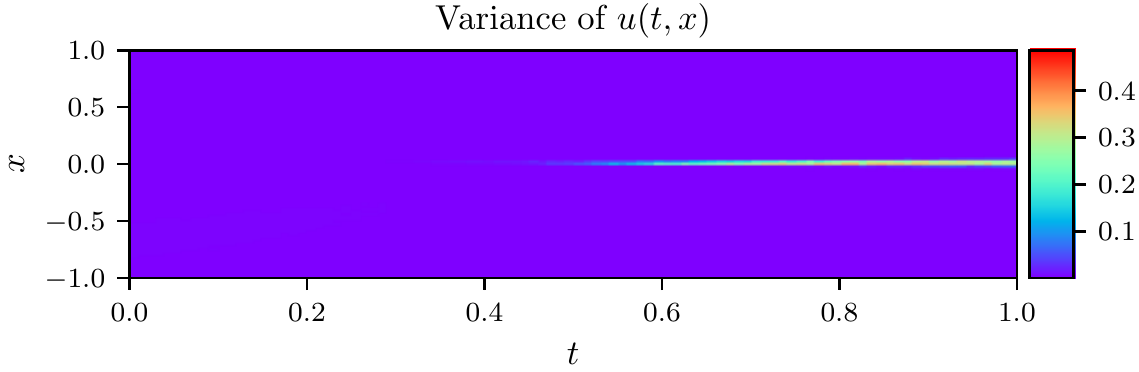}
\caption{{\em Burgers equation with noise-free data:} {\it Top:} Mean of $p_{\theta}(u|x,t,z)$, along with the location of the noisy training data $\{(x_{i}, t_i), u_i\}$, $i=1,\dots,N_u$. {\it Middle:} Prediction and predictive uncertainty at $t=0.25$, $t=0.5$ and $t=0.75$. {\it Bottom:} Variance of $p_{\theta}(u|x,t,z)$.}
\label{fig:Burgers_baseline}
\end{figure}

First we consider a baseline scenario, in which we train our probabilistic model using a data-set comprising of $N_u = \textcolor{black}{1}50$ noisy-free input/output pairs for $u(x,t)$ -- 50 points for the initial condition (see figure \ref{fig:Burgers_initial}(a)) and \textcolor{black}{50} points for each of the domain boundaries -- plus an additional  $N_r = 10,000$ collocation points for enforcing the residual of the Burgers equation using the loss of equation  \ref{eq:PI_loss}. All data points were randomly selected within the bounds given in equation \ref{eq:Burgers}. The result of this experiment is summarized in \ref{fig:Burgers_baseline} where we report the predicted mean solution, as well as the uncertainty associated with this prediction as quantified by two standard deviations of the generative model $p_{\theta}(u|x,t,z)$. As the training data for this case is noise-free, the solution to this problem is deterministic, and the resulting uncertainty captured in $p_{\theta}(u|x,t,z)$ can be viewed as an {\em a-posteriori} error estimate of  the neural network approximation error due to the finite number of training data, which is measured as $\mathcal{E}_{\mathbb{L}_2} = 4.1\cdot 10^{-2}$ in the relative $\mathbb{L}_{2}$ norm. As discussed in \cite{raissi2017physicsI}, a higher approximation accuracy can be achieved by training the generative model using a quasi-Newton optimizer (e.g. L-BFGS \cite{liu1989limited}), however here we chose to use stochastic gradient descent using Adam updates \cite{kingma2014adam} in order to highlight the ability of the proposed method to return uncertainty estimates when the model predictions are not perfectly accurate.

\begin{figure}
\centering
\includegraphics[width=\textwidth]{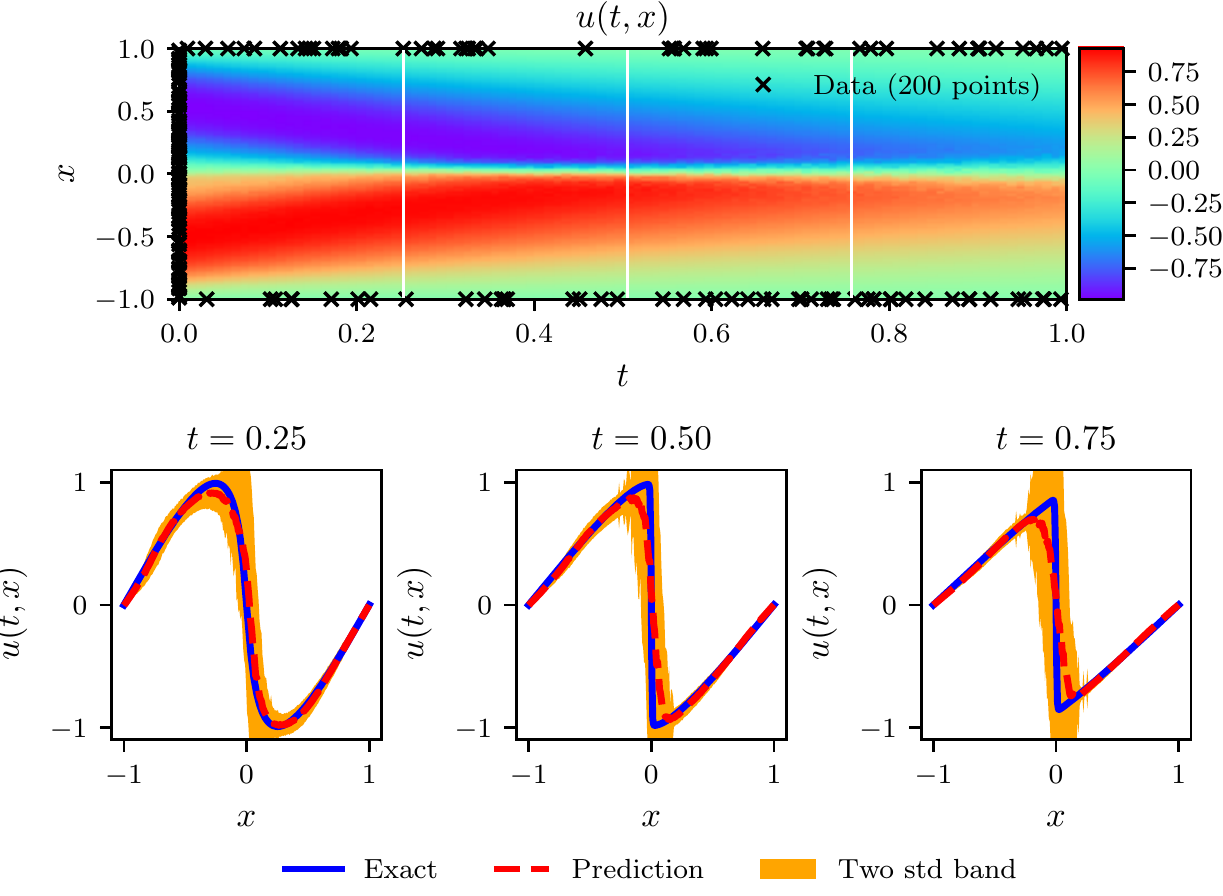}
\includegraphics[width=\textwidth]{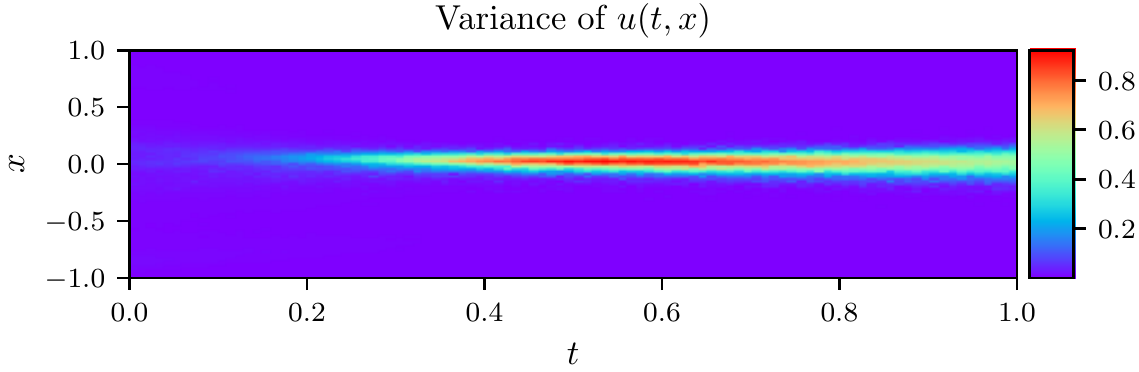}
\caption{{\em Burgers equation with noisy data:} {\it Top:} Mean of $p_{\theta}(u|x,t,z)$, along with the location of the training data $\{(x_{i}, t_i), u_i\}$, $i=1,\dots,N_u$. {\it Middle:} Prediction and predictive uncertainty at $t=0.25$, $t=0.5$ and $t=0.75$. {\it Bottom:} Variance of $p_{\theta}(u|x,t,z)$.}
\label{fig:Burgers_noisy}
\end{figure}

Second, we repeat the same test for a more complicated scenario in which the initial condition has been corrupted by non-additive, non-Gaussian noise as shown in figure \ref{fig:Burgers_initial}(b), where the noise variance is larger around $x=0$, therefore amplifying the effect of uncertainty on the shock formation. Here the neural network architecture as well as the number and location of training points have been kept fixed as described above, but the initial condition is now corrupted as 
\begin{equation}
\label{eq:noisy_data}
u(x, 0) = -\sin(\pi(x+2\delta))+\delta, \qquad \delta = \frac{\epsilon}{\exp(3|x|)}, \qquad \epsilon\sim N(0, 0.1^2). 
\end{equation}
The results of this experiment are summarized in figure \ref{fig:Burgers_noisy}. We observe that the resulting generative model $p_{\theta}(u|x,t,z)$ can effectively capture the uncertainty in the resulting spatio-temporal solution due to the propagation of the input noise process through the complex non-linear dynamics of the Burgers equation. As expected, the uncertainty concentrates around the shock. Although we only plot the first two moments of the solution, we must emphasize that the generative model $p_{\theta}(u|x,t,z)$ provides a complete probabilistic characterization of its non-Gaussian statistics.

In order to further investigate the performance of the proposed methodology for different parameter settings, we have performed a series of comprehensive systematic studies that aim to quantify the sensitivity of the resulting predictions on: (i) the neural network initialization, (ii) the total number of training and collocation points, (iii) the neural network architecture, and (iv) the adversarial training procedure. The results of these systematic studies are provided in \ref{sec:appendix}.

\subsection{Discovery of constitutive laws for flow through porous media}
In our final example, we aim to demonstrate the ability of the proposed methods to discover unknown  constitutive relationships directly from data with quantified uncertainty. To this end, we revisit the Darcy flow example put forth in \cite{tartakovsky2018learning} corresponding to a two-dimensional nonlinear diffusion equation with an unknown state-dependent diffusion coefficient
\begin{equation}
\label{eq:diffusion}
\begin{aligned}
	\nabla_{\bm{x}}\cdot[K(u)\nabla_{\bm{x}} u(\bm{x})] &= 0, \quad\quad \bm{x}=(x_1, x_2)\in \Omega = (0, L_1)\times (0, L_2)\\
	u(\bm{x}) &= u_0, \quad\quad\quad\;\; x_1 = L_1\\
	-K(u)\frac{\partial u(\bm{x})}{\partial x_1} &= q, \quad\quad\quad\quad x_1 = 0\\
	\frac{\partial u(\bm{x})}{\partial x_2} &= 0, \quad\quad\quad\quad x_2 = \{0, L_2\},\\	
\end{aligned}
\end{equation}
where $q = 8.25 \times 10^{-5}$m/s and $u_0 = - 10$m are known boundary conditions. 
In order to benchmark and validate our model predictions we consider a realistic data-set generated using the Subsurface Transport
Over Multiple Phases (STOMP) code \cite{white1995modeling} with the van Genuchten model
\cite{van1980closed} for $K(u)$ which reads as
\begin{equation}
\label{eq:STOMP}
\begin{aligned}
	K(s(u)) &= K_s s^{\frac{1}{2}}[1-(1-s^{\frac{1}{m}})^m]^2\\
	s(u) &= \{1+ [\alpha (u_g - u)]^{\frac{1}{1-m}}\}^{-m},\\
\end{aligned}
\end{equation}
with the following parameter values: $K_s = 8.25 \times 10^{-4}$ m/s, $u_g = 0$, $m = 0.469$, $\alpha = 0.1$, $L_1 = 10$ m and $L_2 = 10$ m.

Our goal is twofold: we aim to construct a physics-informed probabilistic model for $p_{\theta}(u|\bm{x},\bm{z})$, and simultaneously learn the unknown state-dependent diffusion coefficient $K(u)$ directly from data on $u(\bm{x})$ (i.e., we assume no measurements of $K(u)$). To this end, in addition to the three deterministic mappings $f_{\theta}(\bm{x}, \bm{z})$, $q_{\phi}(\bm{x}, u)$, and $T_{\psi}(\bm{x}, u)$ corresponding to the generator, encoder, and discriminator described in section \ref{sec:ADVI}, here we also introduce 
another neural network $f_{\gamma}(u)$ for approximating $K(u)$. The parameters of $f_{\gamma}(u)$ are essentially inherited by the physics-informed residual network $r_{\theta,\gamma}(\bm{x}) := \nabla_{\bm{x}}\cdot[f_{\gamma}(f_{\theta}(\bm{x}, \bm{z}))\nabla_{\bm{x}} f_{\theta}(\bm{x}, \bm{z}))] $ that aims to enforce the residual of equation \ref{eq:diffusion} at the $N_r$ collocation points for any set of latent variables $\bm{z}$. All neural networks are chosen to have 2 hidden layers with 50 neurons per each, and a hyperbolic tangent activation function, while the probabilistic model for $p_{\theta}(u|\bm{x},\bm{z})$ assumes a two dimensional latent space with an isotropic Gaussian prior, i.e., $\bm{z} \sim \mathcal{N}(\bm{0}, \bm{I})$.

By construction, our probabilistic model for $p_{\theta}(u|\bm{x},\bm{z})$ can return predictions of the unknown solution $u(\bm{x})$ with quantified uncertainty.
We can then use this model to propagate uncertainty in our predictions of $K(u)$ via Monte Carlo sampling. Specifically, once the model is trained end-to-end, we can easily generate samples of $u(\bm{x})$ from $p_{\theta}(u|\bm{x},\bm{z})$ and propagate them through $f_{\gamma}(u)$ to obtain a samples for $K(u)$. Essentially this results in an implicit generative model $p_{\theta,\gamma}(k|\bm{x},\bm{z})$ which can fully characterize uncertainty in our predictions of the unknown state-dependent diffusion coefficient.

\begin{figure}
\centering
\includegraphics[width=\textwidth]{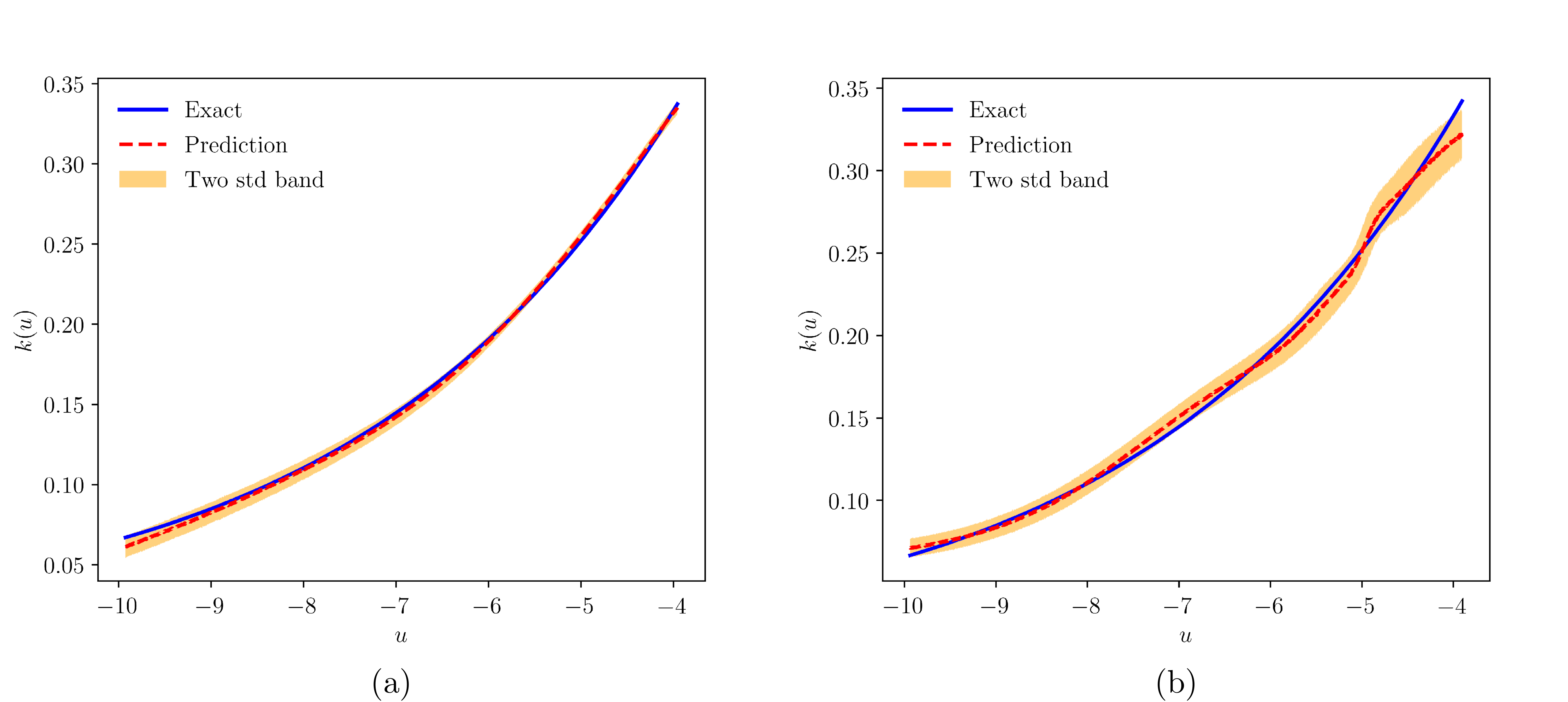}
\caption{Prediction with quantified uncertainty of unknown state-dependent diffusion coefficient $K(u)$ compared against the reference solution obtained from the Subsurface Transport
Over Multiple Phases (STOMP) code \cite{white1995modeling} with the van Genuchten model
\cite{van1980closed} (see equation \ref{eq:STOMP}). (a) Noise-free training data for $u(\bm{x})$. (b) Noisy training data for $u(\bm{x})$ with noise level of $5\%$.}
\label{fig:k_u}
\end{figure}

\begin{figure}
\centering
\includegraphics[width=\textwidth]{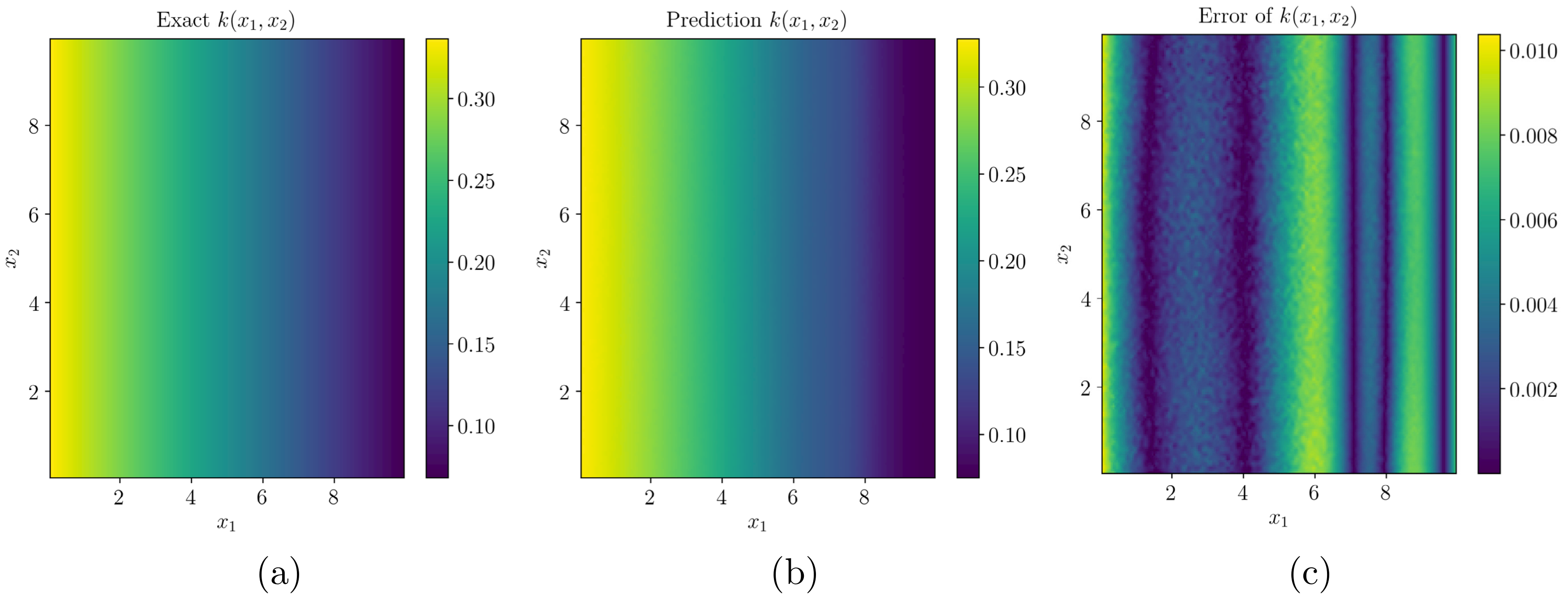}
\caption{Prediction of unknown state-dependent diffusion coefficient $K(u(x_1,x_2))$. (a) Reference solution obtained from the Subsurface Transport
Over Multiple Phases (STOMP) code \cite{white1995modeling} with the van Genuchten model
\cite{van1980closed} (see equation \ref{eq:STOMP}). (b) Predictive mean of the generative model $p_{\theta,\gamma}(k|\bm{x},\bm{z})$ trained on noise-free data for $u(\bm{x})$. (c) Absolute point-wise prediction error.}
\label{fig:k_x}
\end{figure}

Here we also have considered two cases corresponding to noise-free training data and noisy data corrupted by $5\%$ Gaussian uncorrelated noise. In the \textcolor{black}{noise-free} case we used $N_u = 600$ scattered measurements of the unknown solution $u(\bm{x})$ -- 200 inside the domain $\Omega$ and 100 on each one of the four boundaries -- and total number of $N_r = 10,000$ randomly selected collocation points inside the domain for penalizing the residual of equation \ref{eq:diffusion}. \textcolor{black}{For the noisy case we chose $N_u = 1,400$ scattered measurements of the unknown solution $u(\bm{x})$ -- 1,000 inside the domain $\Omega$ and 100 on each one of the four boundaries -- while still keeping $N_r = 10,000$ collocation points.} Figure \ref{fig:k_u} summarizes the results for both cases by showing the predictive mean and two standard deviations of the corresponding generative model $p_{\theta,\gamma}(k|\bm{x},\bm{z})$, against the reference (deterministic) solution obtained from the Subsurface Transport Over Multiple Phases (STOMP) code \cite{white1995modeling} with the van Genuchten model \cite{van1980closed} (see equation \ref{eq:STOMP}). Evidently, the generative model is able to recover a sensible prediction for the unknown state-dependent diffusion coefficient with quantitative uncertainty, even when the training data on $u(\bm{x})$ is heavily corrupted by noise. Moreover, notice that  $k(u)$ implicitly depends on the spatial coordinates $\bm{x}=(x_1, x_2)$. In figure \ref{fig:k_x} we present the resulting prediction for $K(u(x_1,x_2))$ corresponding to the noise-free case, against the reference solution, as well as their point-wise absolute error.

Again we must emphasize that these predictions are obtained without ever observing any data on $K(u)$, while they are accompanied by quantitative estimates that jointly characterize the uncertainty due to noise in the training data for $u(\bm{x})$, and the underlying approximation error of the neural networks. This theme of consistently inferring correlated continuous quantities of interest from a small set of measurements by leveraging the underlying laws of physics is a great example of the exciting capabilities that physics informed machine learning has to offer.

\section{Conclusions}

We presented a class of  probabilistic physics-informed neural networks that are capable of approximating arbitrary conditional probability densities, while being constrained to generate samples that approximately satisfy given partial differential equations. Moreover, we have derived a flexible regularized adversarial inference framework that enables the end-to-end training of such models directly from noisy and incomplete measurements. Uncertainty in the system inputs and/or outputs is captured through a set of latent variables that are relevant to the underlying physics, and could possibly open new directions for probabilistic model-order reduction of complex systems. These developments allow us to perform probabilistic computations for uncertain systems, train deep generative models in small data regimes, handle complex noise processes, and seamlessly carry out uncertainty propagation studies for physical systems without the need for repeated evaluation of experiments and numerical simulations.

Although the proposed adversarial inference framework provides great flexibility for performing probabilistic computations and approximating arbitrarily complex and high-dimensional probability distributions, it relies on carefully tuning the interplay between the generator and discriminator networks. This is a known limitation of adversarial algorithms, and, although several works have led to improvements \cite{tolstikhin2017wasserstein, arjovsky2017wasserstein}, it still largely remains an open research problem. An alternative path for enhancing the robustness of the inference procedure, while not compromising its ability to handle complex probability distributions, comes through the use of invertible transformations and flow-based generative models \cite{rezende2015variational, kingma2018glow}. 
Future work will examine the applications of such models in the context of physics-informed neural networks with the goal of robustifying the proposed methods and scaling them to more realistic systems.

\section*{Acknowledgements}
This work received support from the US Department of Energy under the Advanced Scientific Computing Research program (grant DE-SC0019116) and the Defense Advanced Research Projects Agency under the Physics of Artificial Intelligence program. We would also like to thank Dr. Alexandre Tartakovsky from the Pacific Nortwest National Laboratory for providing the Darcy flow data-set.

\input{appendix.tex}

\end{document}

%% file: appendix.tex
\appendix

\section{Sensitivity studies}
\label{sec:appendix}

Here we provide results on a series of comprehensive systematic studies that aim to quantify the sensitivity of the resulting predictions on: (i) the neural network initialization, (ii) the total number of training and collocation points, (iii) the neural network architecture, and (iv) the adversarial training procedure. In all cases we have used the non-linear Burgers defined in section \ref{sec:burgers} as a prototype problem.

\subsection{Sensitivity with respect to the neural network initialization}

In order to quantify the sensitivity of the proposed methods with respect to the initialization of the neural networks, we have considered a noise-free data set comprising of $N_u = 150$ and $N_r = 10000$ training and collocation points, respectively, and fixed the architecture for generator neural networks to include 4 hidden layers with 50 neurons each and discriminator neural networks to include 3 hidden layers with 50 neurons each
, and a hyperbolic tangent activation function. Then we have trained an ensemble of 15 cases all starting from a normal Xavier initialization \citep{glorot2010understanding} for all network weights (with a randomized seed), and a zero initialization for all bias parameters. In table \ref{tab:sens_t1} we report the relative error between the predicted mean solution and the known exact solution for this problem for all 15 randomized trials using at set of $25600$ randomly selected test points. Evidently, our results are robust with respect to the the neural network initialization as in all cases the stochastic gradient descent training procedure converged roughly to the same solution. We can summarize this result by reporting the mean and the standard deviation of the relative $\mathbb{L}_2$ error as 
\[
\hat{\mathcal{L}_2} \in [\mu_L-\sigma_L, \mu_L+\sigma_L] = [4.7\times 10^{-2} - 1.3\times 10^{-2}, 4.7\times 10^{-2} + 1.3\times 10^{-2}].
\]

\begin{table}
\centering
\begin{tabular}{|c|c|c|c|c|c|}
\hline
\multicolumn{5}{|c|}{Relative $\mathcal{L}_2$ error}\\ 
\hline
4.1e-02& 7.9e-02& 4.4e-02& 4.0e-02& 3.8e-02\\
\hline
3.2e-02& 5.7e-02& 4.7e-02& 6.5e-02& 4.0e-02\\
\hline
3.5e-02& 3.5e-02& 6.4e-02& 4.0e-02& 4.9e-02\\
\hline
\end{tabular}
\caption{Relative $\mathbb{L}_2$ prediction error for different neural network initializations using a randomized seed.}
\label{tab:sens_t1}
\end{table}

\subsection{Sensitivity with respect to the total number of training and collocation points}\label{B2}

In this study our goal is to quantify the sensitivity of our predictions with respect to the total number of training and collocation points $N_u$ and $N_r$, respectively. As before, we have considered noise-free data sets, and fixed the architecture for generator neural networks to include 4 hidden layers with 50 neurons each and discriminator neural networks to include 3 hidden layers with 50 neurons each, a hyperbolic tangent activation function, and a normal Xavier initialization \citep{glorot2010understanding} for all network weights and zero initialization for all network biases. The results of this study are summarized in table \ref{tab:sens_t2}, indicating that as the number of collocation points are increased, a more accurate prediction is obtained. This observation is in agreement with the original results of Raissi {\em et. al.} \citep{raissi2017physics1,raissi2017physics2} for deterministic physics-informed neural networks, indicating the role of the residual loss as an effective regularization mechanism for training deep generative models in small data regimes.

\begin{table}
\centering
\begin{tabular}{|c|c|c|c|c|c|c|c|c|c|}
\hline
\diagbox{$N_u$}{$N_r$} & 10& 100& 250& 500& 1000& 5000& 10000\\ 
\hline
60  & 9.3e-01& 5.6e-01& 4.8e-01& 5.0e-02& 1.9e-01& 5.0e-02& 5.1e-02\\
\hline
90  & 5.8e-01& 5.3e-01& 3.5e-01& 1.5e-01& 4.9e-02& 1.0e-01& 5.8e-02\\
\hline
150 & 6.7e-01& 1.4e-01& 3.0e-01& 3.6e-02& 4.9e-02& 1.2e-01& 4.7e-02\\
\hline
\end{tabular}
\caption{Relative $\mathbb{L}_2$ prediction error for different number of training and collocation points $N_u$ and $N_r$, respectively.}
\label{tab:sens_t2}
\end{table}

\subsection{Sensitivity with respect to the neural network architecture}

In this study we aim to quantify the sensitivity of our predictions with respect to the architecture of the neural networks that parametrize the generator, the discriminator, and the encoder. Here we have fixed the number of noise-free training data to $N_u = 150$ and $N_r =10000$, and we kept the number of layers for discriminator to always be one less than the number of layers for generator (e.g., if the number of layers for generator is two then the number of layers for discriminator is one, etc.). 
In all cases, we have used a hyperbolic tangent non-linearity and a normal Xavier initialization \citep{glorot2010understanding}. In table \ref{tab:sens_t3} we report the relative $\mathbb{L}_2$ prediction error for different feed-forward architectures for the generator, discriminator, and encoder (i.e., different number of layers and number of nodes in each layer). The general trend suggests that as the neural network capacity is increased we obtain more accurate predictions, indicating that our physics-informed constraint on the PDE residual can effectively regularize the training process and safe-guard against over-fitting. We note number of neurons in each layer as $N_{n}$ and number of layers for generator (encoder) as $N_g$.

\begin{table}
\centering
\begin{tabular}{|c|c|c|c|}
\hline
\diagbox{$N_g$}{$N_n$} & 20& 50& 100\\ 
\hline
2 &    4.2e-01& 3.8e-01& 5.7e-01 \\
\hline
3 &    6.5e-02& 3.5e-02& 2.1e-02 \\
\hline
4 &    9.3e-02& 4.7e-02& 5.4e-02 \\
\hline
\end{tabular}
\caption{Relative $\mathbb{L}_2$ prediction error for different feed-forward architectures for the generator, encoder, and the discriminator. The total number of layers of the latter was always chosen to be one less than the number of layers for generator.}
\label{tab:sens_t3}
\end{table}

\subsection{Sensitivity with respect to the adversarial training procedure}

Finally, we test the sensitivity with respect to the adversarial training process. To this end, we have fixed the number of noise-free training data to $N_u = 150$ and $N_r =10000$, and the neural network architecture to be the same as \ref{B2}, and we vary the total number of training steps for the generator $K_g$ and the discriminator $K_d$ within each stochastic gradient descent iteration. The results of this study are presented in table \ref{tab:sens_t4} where we report the relative $\mathbb{L}_2$ prediction error. These results reveal the high sensitivity of the training dynamics on the interplay between the generator and discriminator networks, and pinpoint on the well known peculiarity of adversarial inference procedures which require a careful tuning of 
$K_g$ and $K_d$ for achieving stable performance in practice.

\begin{table}
\centering
\begin{tabular}{|c|c|c|c|}
\hline
\diagbox{$K_g$}{$K_d$} & 1& 2& 5\\ 
\hline
1 &    3.5e-01& 5.0e-01& 1.5e+00 \\
\hline
2 &    4.3e-02& 3.2e-01& 5.4e-01 \\
\hline
5 &    4.7e-02& 2.3e-01& 7.0e-01 \\
\hline
\end{tabular}
\caption{Relative $\mathbb{L}_2$ error with different number of training for generator and discriminator in each epoch.}
\label{tab:sens_t4}
\end{table}